\documentclass[fleqn,10pt]{wlscirep}
\usepackage[utf8]{inputenc}
\usepackage[T1]{fontenc}
\usepackage{multirow}
\usepackage{wasysym}
\usepackage{amssymb}
\usepackage{longtable}
\usepackage[nodisplayskipstretch]{setspace}
\usepackage{amsmath}
\usepackage{nccmath}
\title{InForecaster: Forecasting Influenza Hemagglutinin Mutations Through the Lens of Anomaly Detection
}

\author[1]{Ali Garjani}
\author[1]{Atoosa Malemir Chegini}
\author[1]{Mohammadreza Salehi}
\author[2]{Alireza Tabibzadeh}
\author[2]{Parastoo Yousefi}
\author[2]{Mohammad Hossein Razizadeh}
\author[3]{Moein Esghaei}
\author[2]{Maryam Esghaei}
\author[1,*]{Mohammad Hossein Rohban}

\affil[1]{Department of Computer Engineering, Sharif University of Technology, Tehran, Iran}
\affil[2]{Department of Virology, School of Medicine, Iran University of Medical Sciences, Tehran, Iran}
\affil[3]{Cognitive Neuroscience Laboratory, German Primate Center, Leibniz Institute for Primate Research, Goettingen, Germany}

\affil[*]{Corresponding Author's email: rohban@sharif.edu}


\begin{abstract}
The influenza virus hemagglutinin is an important part of the virus attachment to the host cells. The hemagglutinin proteins are one of the genetic regions of the virus with a high potential for mutations. Due to the importance of predicting mutations in producing effective and low-cost vaccines, solutions that attempt to approach this problem have recently gained a significant attention. A historical record of mutations have been used to train  predictive models in such solutions. However, the imbalance between mutations and the preserved proteins is a big challenge for the development of such models that needs to be addressed. Here, we propose to tackle this challenge through anomaly detection (AD).  AD is a well-established field in Machine Learning (ML) that tries to distinguish unseen anomalies from the normal patterns using only normal training samples. By considering mutations as the anomalous behavior, we could benefit existing rich solutions in this field that have emerged recently. Such methods also fit the problem setup of extreme imbalance between the number of unmutated vs. mutated training samples. Motivated by this formulation, our method tries to find a compact representation for unmutated samples while forcing anomalies to be separated from the normal ones. This helps the model to learn a shared unique representation between normal training samples as much as possible, which improves the discernibility and detectability of mutated samples from the unmutated ones at the test time. We conduct a large number of experiments on four publicly available datasets, consisting of 3 different hemagglutinin protein datasets, and one SARS-CoV-2 dataset,  and show the effectiveness of our method through different standard criteria.  

\end{abstract}
\begin{document}

\flushbottom
\maketitle
%
\thispagestyle{empty}



\section*{Introduction}

The influenza virus infection is mostly presented as a self-limited respiratory infection in immunocompetent people. However, influenza viruses could lead to a life-threateing infection in the eldery and other risk group patients. Hemagglutinin is a glycoprotein located on the surface of influenza viruses and acts as an attaching ligand to the host cells and inserting the virus into the cells. To escape from immune responses, the virus can alter the antigenic features of the hemagglutinin (HA) protein by point mutations. This phenomenon is known as antigenic drifts. Mutations in the genes of influenza viruses could cause antigenic drift by changing the HA protein structure \cite{bush1999predicting, banning2005influenza, simonsen1997impact, webster1992evolution}.  This results in a new strain of the virus  that is not effectively recognized by the immune system and makes the virus spread easily and cause epidemic. Influenza viruses are classified in the {\it Orthomyxoviridae} family. In this family, there are three important human pathogens, including {\it Alphainfluenzavirus}, {\it Betainfluenzavirus}, and {\it Gammainfluenzavirus}. Influenza A virus is the only member of {\it Alphainfluenzavirus}, and is an important human pathogen due to its wide host range and the higher rate of drift and shift mutations \cite{bedford2015global, chen2007exploration, bodewes2013recurring}.

HA is the main part of the virus attachment to the host cell receptor. The globular head domain of HA, which is critical for neutralizing antibody generation by the host immune system, is one of the most potent genomic locations for mutation. Influenza A viruses are divided into two different groups based on the globular head domain. The HA 1, 2, 5, 6, 9, 11, 12, 13, 16, and 18 types are placed in one group and types 3, 4, 7, 10, 14, and 15 are considered as members of the second group of HAs \cite{krammer2019human}. The Cb, Ca, Sb, and Sa are four important antigenic sites in the H1 domain of HA \cite{luoh1992hemagglutinin, caton1982antigenic, brownlee2001predicted}. The amino acid residues number 143, 156, 158, 190, 193, and 197 are the most important residues for evolutionary and antigenic features of HA1 \cite{shen2009evolutionary}. The role of HA1 mutations in the adequacy of influenza vaccination has made WHO Collaborating Centers and Vaccines and Related Biological Products Advisory Committee (VRBPAC)\cite{buckland2015development} responsible for functional monitoring, reports, and decision for new season vaccines. Despite this delicate process, there are shortages and some strain mismatches between the vaccine strains and circulating strains \cite{ampofo2015strengthening, lin2015optimisation}. 

In the current study, we also evaluated the SARS-CoV-2 Spike mutations as an extra evaluation and a demo for the future consideration in this field. The SARS-CoV-2 (severe acute respiratory syndrome coronavirus 2) is the etiological agent for the COVID-19 (Coronavirus disease-2019) pandemic. The SARS-CoV-2 is a Betacoronavirus and a member of the sarbecoviruses sublineage \cite{Tabibzadeh2020-pe}. The virus genome is an ssRNA (Single strand RNA) of 34kb in length. SARS-CoV-2 contains different genes including ORF1a/b, Spike (S), Envelope (E), Membrane (M), Nucleoprotein (N), and accessory ORFs \cite{Kumar2020-cg}. The virus binding into cells by using the S protein attachment to the cellular receptor ACE-2 (Angiotensin-converting enzyme 2) \cite{Bourgonje2020-ou}. The S protein is the most important antigenic part of the virus \cite{Xiaojie2020-fk}. 

In recent years, Artificial Intelligence (AI) algorithms have achieved human or even super-human performance on different tasks such as image classification\cite{tsipras2020imagenet}, text classification\cite{kowsari2019text}, action recognition\cite{zhang2019comprehensive}, etc. Anomaly Detection (AD) is a sub-domain of AI that is responsible for learning a normal representation space, and detecting anomalous samples at the test time by exploiting the learned representation. Due to different challenges in labeling of the anomalous samples, such as the high cost or rareness of such samples, most methods in this domain only use normal samples for the training. This is called the unsupervised AD. Alternatively, one may use a very limited number of labeled anomalous samples in the training process, which is called the semi-supervised AD\cite{chalapathy2019deep}.

Unsupervised\cite{ruff2018deep, salehi2020arae, akcay2018ganomaly, schlegl2017unsupervised, goyal2020drocc} and semi-supervised \cite{ruff2019deep, ruff2020rethinking} anomaly detection methods have recently achieved satisfactory results on a variety of domains such as image, text, time-series, and video.  Deep Semi-Supervised Anomaly Detection (DeepSAD)\cite{ruff2019deep}, as a recently proposed semi-supervised AD method, made clear that semi-supervised anomaly detectors are significantly superior compared to the supervised training classification algorithms, specifically when the training dataset is complex, and the number of normal samples is much higher than the anomalous ones. This is because anomaly detectors attempt to find a compact representation space for the normal samples while maximizing the margin that exists between normal and abnormal ones. This helps them to learn the most general and unique features of the normal samples, and not rely overly on the contrast that exists between normal and anomalous samples to classify them.

Since in the mutation prediction tasks the number of unmutated samples is much higher than the mutated ones, the problem can be formulated as an anomaly detection task. In this formulation, unmutated and mutated samples are considered as normal and anomalous samples, respectively. The benefits of this approach are two-fold. Firstly, a semantically meaningful representation could be learned even with a small number of training samples, which makes generalization to unseen test time samples possible. Secondly, as the finding and labeling procedure of mutated viruses is an expensive and time-consuming process, anomaly detectors could work fine with, or without a limited number of anomalous or mutated, training samples\cite{chalapathy2019deep}.

\begin{figure*}
  \centering
  \includegraphics[width=0.85\linewidth, scale=0.1]{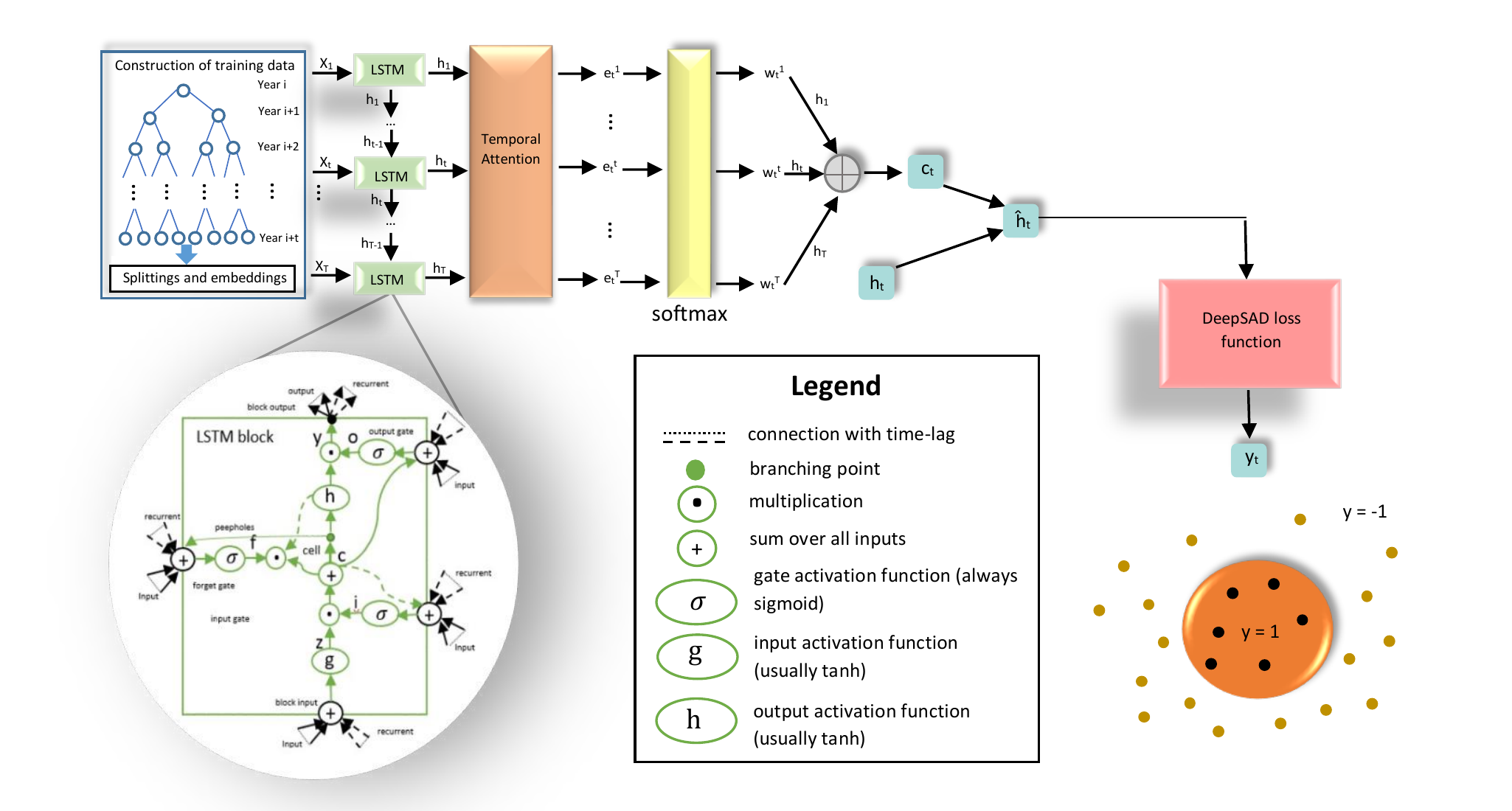}
  \caption{The overall architecture of our method. First, the raw data is processed and the output $(X^{t}_1, X^{t}_2,\dots, X^{t}_n)$ is prepared at the time step $t$, where $n$ is the embedding dimensions, $t$ denotes the time. After the pre-processing phase, LSTM cells are used to produce hidden states, $h_i$, for each time point $t$. Then, the attention function takes $h_i$ and the cell state $s_{t-1}$, and outputs $e^{i}_{t}$. Next, by using a softmax function, the weights $w^{i}_{t}$'s are produced. The weighted sum of the hidden states, $h_{i}$'s, is obtained by using the mentioned weights. The output of this weighted sum, $c_t$, and the hidden state $h_{t}$ will then be used to produce the encoded vector $\hat{h}_t$. At the last step, DeepSAD loss function is applied to $\hat{h}_t$ to decide whether the input data is in-class (normal) 1 or out-class (anomaly) -1.}
  \label{fig:1}
\end{figure*}

By this motivation, we propose the first anomaly detection framework for predicting virus mutations. We use the Long Short-Term Memory (LSTM)\cite{hochreiter1997long} neural network in combination with the Deep Semi-Supervised Anomaly Detection (DeepSAD) loss\cite{ruff2019deep} to not only learn long-term input dependencies, but also to find a semantic representation space for the mutated and unmutated training samples. Figure \ref{fig:1} shows the overall architecture of the proposed method.  We conduct extensive experiments to show the effectiveness of our method in improving the average recall, F1-score, precision, and Area Under the Curve (AUC) for three different publicly available Influenza datasets.

\section*{Background}
For the sake of clarity, we discuss some of the important prerequisites from deep learning literature in this section. At first, some Recurrent Neural Network architectures, such as LSTMs\cite{hochreiter1997long}, are discussed. Then, a brief introduction about the anomaly detection methods is presented.

\textbf{Recurrent Neural Networks (RNN):}
RNNs are broadly used to model the data sequential dependencies, where the sequence could be formed based on the temporal or spatial arrangements. Initial architectures of RNNs, such as the vanilla RNN, suffer from memorizing long-term as well as short-term dependencies. To address this issue, alternative architectures, such as LSTM\cite{hochreiter1997long} networks, bi-directional RNNs\cite{schuster1997bidirectional}, and gated recurrent units\cite{chung2014empirical} GRU's have been introduced. All these approaches attempt to summarize previous inputs into their hidden state that is updated in each time step $t$. The mentioned information is regulated using some parameters or gates. For instance, the LSTM network consists LSTM cells. Each cell contains a state, $h_t$, and memory, $s_t$. These two are updated based on three different gates that are called \emph{input gate}, $i_t$, \emph{forget gate}, $f_t$, and \emph{output gate}, $o_t$. The input gate selects some of the memory dimensions to modify (Eq. \ref{eq:2}). The forget gate decides which memory cell dimensions should be ignored in the next time step (Eq. \ref{eq:1}). The output gate decides which dimensions of the memory should be transferred to the state (Eq. \ref{eq:3}). The cell and state vectors are updated based on these gates and activation values that are produced through the tanh activation (Eqs. \ref{eq:4}, \ref{eq:5}). Specifically, the memory constitutes previous memory dimensions that are not forgotten, plus the input activation values that the input gate selects. Finally, the state constitutes memory activation values that are selected by the output gate.

Note that a sigmoid activation function is used in the gates to map the gate outputs between zero and one, which models the selection, i.e. gate output of 1 represents complete selection of an embedding, and the 0 value corresponds to a complete non-selection. A tanh activation function is used in the cell and state update rules to produce activation values that are between -1 and 1.  

\begin{ceqn}
\begin{align}
f^{}_{t} = \sigma (W^{}_{f} [h^{}_{t-1} ; x^{}_{t}] + b^{}_{f} ) \label{eq:1}
\end{align}
\addtolength\abovedisplayskip{-1.8\baselineskip}
\addtolength\belowdisplayskip{0\baselineskip}
\begin{align}
i^{}_{t} = \sigma (W^{}_{i} [h^{}_{t-1} ; x^{}_{t}] + b^{}_{i} ) \label{eq:2}
\end{align}
\addtolength\abovedisplayskip{0.2\baselineskip}
\addtolength\belowdisplayskip{0\baselineskip}
\begin{align}
o^{}_{t} = \sigma (W^{}_{o} [h^{}_{t-1} ; x^{}_{t}] + b^{}_{o} ) \label{eq:3}
\end{align}
\addtolength\abovedisplayskip{0\baselineskip}
\addtolength\belowdisplayskip{0\baselineskip}
\begin{align}
s^{}_{t} = f^{}_{t} \odot s^{}_{t-1} + i^{}_{t} \odot tanh(W^{}_{s}[h^{}_{t-1} ; x^{}_{t}] + b^{}_{s}) 
\label{eq:4}
\end{align}
\addtolength\abovedisplayskip{0\baselineskip}
\addtolength\belowdisplayskip{0\baselineskip}
\begin{align}
h^{}_{t} = o^{}_{t} \odot tanh(s^{}_{t}) \label{eq:5}
\end{align}
\end{ceqn}

\addtolength\abovedisplayskip{0.5\baselineskip}
\addtolength\belowdisplayskip{0\baselineskip}

Despite huge efforts on making different LSTM architectures to improve its performance, no architecture has been proposed yet that is generally better than the original one\cite{greff2016lstm}. Therefore, our proposed method is based on the LSTM networks with some improvements on its ability to maintain long-term information and interpretability.

\textbf{Anomaly Detection:} As mentioned before, anomaly detection is a sub-branch of artificial intelligence seeking to solve one-class classification problems. One-class methods only access to the labels of one category of a dataset, called the normal class \cite{salehi2021unified}. These methods then seek to design a classifier that can distinguish the normal class vs. the unseen classes, which is also referred to as anomaly classes.
For instance, in the mutation prediction problems, the anomaly detection method assumes access to only unmutated samples. This setup could be adopted for reasons such as the large cost of the data gathering process from both kinds of mutated and unmutated classes, or even the impossibility of gathering all kinds of mutations in our training dataset. Such issues make the classification setup ineffective, as the classifier may get biased towards accurate prediction of only known mutations that are reflected in the training set. 
Deep Support Vector Data Description (DSVDD)\cite{ruff2018deep} is one of the basic anomaly detection methods that is trained in an unsupervised manner. It tries to find a latent space and the most compact hyper-sphere that contains the normal training samples in this space. The pre-assumption of DSVDD is that anomalous samples layout of the circle in contrast to normal ones, which could make them detectable. Recently, Chong et al. \cite{chong2020simple} has shown the vulnerability of this method to the mode collapse, i.e. convergence of all data points to a single point in the latent space, due to its unsupervised training process. To alleviate this issue, DSAD\cite{ruff2019deep} suggest to use a limited number of anomalous training samples to train DSVDD and achieved satisfying results. It has shown that a limited number of anomalous samples is enough to not only prevent the mode collapse but also enhance the supervised classifiers accuracy, specifically when training samples are significantly imbalanced toward the normal class.

From a different point of view, autoencoder(AE) is another dominant framework that is used in the field. Owing to their unsupervised training process, they are intriguing for formulating anomaly detection problems based on their abilities. They are trained on normal training samples by this pre-assumption that they would be reconstructed well at the test time compared to the abnormal inputs. As the primary versions of autoencoders have shown deficiencies in their performance when the training dataset becomes complex, some variants of AE-based methods have been proposed based on generative adversarial networks~\cite{sabokrou2018adversarially, zaheer2020old, perera2019ocgan}, adversarial robust training~\cite{salehi2020arae}, or self-supervised learning methods~\cite{salehi2020puzzle}.

\section*{Experiments and Results}

\subsection*{Dataset}
For the experiments on HA, we used the dataset provided in Tempel\cite{yin2020tempel}. The data is hosted at \url{https://drive.google.com/drive/folders/1-pJGBsVfIqCEizetTQe43OQJvkmhocdW}. This dataset includes the influenza subtypes H1N1, H3N2, and H5N1, which have the sequence length of 566, 566, and 568, respectively. The number of available HA sequences in each year for all three subtypes is provided in table \ref{tab:sample_num}. 

\begin{table}[!tbh]
\centering

\caption{Number of HA samples for each subtypes from 2001 to 2016}
\label{tab:sample_num}
\begin{tabular}{lllllllll}
\hline
Year & 2001 & 2002 & 2003 & 2004 & 2005 & 2006 & 2007 & 2008 \\ \hline
H1N1 & 77   & 40   & 77   & 60   & 92   & 139  & 307  & 322  \\
H3N2 & 53   & 135  & 215  & 180  & 204  & 132  & 285  & 189  \\
H5N1 & 21   & 26   & 45   & 117  & 169  & 231  & 291  & 171  \\ \hline
Year & 2009 & 2010 & 2011 & 2012 & 2013 & 2014 & 2015 & 2016 \\ \hline
H1N1 & 2517 & 957  & 794  & 586  & 715  & 499  & 391  & 687  \\
H3N2 & 340  & 411  & 577  & 868  & 704  & 876  & 1036 & 932  \\
H5N1 & 150  & 169  & 188  & 134  & 166  & 143  & 135  & 17   \\ \hline
\end{tabular}
\end{table}

For the experiments on SARS-CoV-2, we selected the RBD domain of spike nucleotides sequences, with the length of 1273, in United Kingdom from the January 1, 2020 to the last day of the December 2020. The number of the available sequences in each month is provided in table \ref{tab:covid_num}. 

The studies are performed in accordance with the Declaration of Helsinki, and are carried out in accordance with the relevant guidelines and regulations.

\begin{table}[!tbh]
\centering

\caption{Number of sequence samples from January 2020 to December 2020}
\label{tab:covid_num}
\begin{tabular}{lllllllllllll}
\hline
Month & Jan & Feb & Mar & Apr & May & Jun & Jul & Aug & Sep & Oct & Nov & Dec \\ \hline
SARS-CoV-2 & 3   & 116   & 900   & 1500   & 1000   & 960  & 986  & 1031 & 800 & 1146 & 1416 & 900  \\
\hline
\end{tabular}
\end{table}

\subsection*{Implementation}
We use PyTorch and Scikit-learn in our implementations. For the experiments that are performed on H1N1, H3N2, and H5N1, similar to the previously proposed method Tempel\cite{yin2020tempel}, the first 1000 samples per year are chosen for training and testing. Then, the first 80$\%$ of the samples in each year are selected as the training set, and the remaining 20$\%$ are set as the test set. For our model, since validation is also needed for finding the best threshold, the first 10$\%$ of the training set is used for validation, and the remaining 90$\%$ for training. We use a batch size of 256, a learning rate of 0.001, and gradient descent for the  optimization. Also, similar to Tempel, hidden layer size and dropout percentage is set to 128 and 0.5, respectively. As figure S3 shows, validation curves can be used as a good hint to stop the training process on the 50th epoch, in which the model has converged. Therefore, we have trained our model for 50 training epochs.

\subsection*{The Bioinformatics pipeline for the suggested amino acid alteration locations}

All of the amino acid alteration locations that are suggested by the algorithm are evaluated by a simple alignment analysis. After the model training,  possible amino acid alteration locations for 2016 influenza H1N1, H3N2, and H5N1 are listed based on the highest recall and precision. The influenza virus sequences for 2016 are obtained from the NCBI influenza database. A random 100 full-length samples from 2016 influenza circulating strains are used as a sample for amino acid alteration locations evaluation. In addition, the alterations are evaluated based on the suggested vaccine strains for the 2016-2017 season (stains include: A/California/7/2009 (H1N1)pdm09-like virus, A/Hong Kong/4801/2014 (H3N2)-like virus and B/Brisbane/60/2008-like virus (B/Victoria lineage)) \cite{cdc}. All of the sampling influenza sequences from 2016 are aligned, and the alignment and amino acid locations are visualized in CLC Workbench \cite{CLC}. Major epitopes are marked based on the previous studies for H1N1 \cite{de2012molecular} and H3N2 \cite{zost2019identification}. The H5N1 influenza was not reported for the human host during 2016, hence for the mutation assessment for this dataset, we use all of the hosts during 2016 and mostly avians.

\subsection*{Baselines}
In this section, we compare our method with other recently proposed anomaly detection (AD) methods that can be easily adapted to our task. Although some approaches such as Golan et al. \cite{golan2018deep} and Bergman et al. \cite{bergman2020classification} achieve top performance in anomaly detection problems, they use self-supervised learning methods that are specialized for the image processing tasks. Consequently, they reach weak performance on our datasets. Table \ref{tab:2} includes the performance of Bergman et al., called GOAD, compared to the other methods. The parameter $T$ represents the time-series sequence length ending in the year 2016. For the sake of equality in our comparisons, we have chosen $T$ to be 5, 10, and 15, similar to Tempel \cite{yin2020tempel}. Besides, repeating experiments for different $T$ values gives a more comprehensive intuition on the sensitivity of the methods.

Some AD algorithms are not domain-specific such as autoencoder-based (AE) approaches. We have reported the  performance of ARAE\cite{salehi2020arae} as a stable, domain agnostic, and high performance approach. As ARAE is trained in a fully unsupervised training manner, using only unmutated training samples, its results are not competitive with the semi-supervised learning methods. Besides, AEs are not effective when facing complex datasets. 

Moreover, we report the performance of a similar semi-supervised anomaly detection method that has been proposed recently, ESAD \cite{huang2020esad}. It uses an AE-Based approach, but employs both kinds of negative (unmutated) and positive (mutated) samples in its loss function. Finally, the performance of the original vanilla DeepSAD method, without our proposed modifications in Eq. \ref{deep_sad_loss}, is reported in table \ref{tab:2}.

\begin{table}[!tbh]
\centering

\caption{Precision, recall and F1-score on H1N1, H3N2 and H5N1 datasets for $T \in \{5,10,15\}$. Results are for four different methods including ARAE, GOAD, ESAD and original DeepSAD.}
\label{tab:2}
\begin{tabular}{@{}lllllllllllll@{}}
  \toprule
  Dataset & Model &\multicolumn{3}{c}{Precision} & & \multicolumn{3}{c}{Recall} & &\multicolumn{3}{c}{F1-score} \\
  \midrule
   & & T=5 & T=10 & T=15 & & T=5 & T=10 & T=15 & & T=5 & T=10 & T=15 \\
  \cmidrule{3-5} \cmidrule{7-9} \cmidrule{11-13}
  \multirow{3}{*}{H1N1} & ARAE & 0.162 & 0.171 & 0.181 & & 0.472 & 0.501 & 0.532 & & 0.241& 0.255& 0.270\\
  & GOAD & 0.124 & 0.132 & 0.190 & & 0.361 & 0.391 & 0.555 & & 0.185& 0.197& 0.283\\
  & ESAD & 0.384 & 0.406 & 0.419 & & 0.570 & 0.573 & 0.589 & & 0.459& 0.475& 0.490\\
  & DeepSAD & 0.294 & 0.318 & 0.322 & & 0.497 & 0.477 & 0.478 & & 0.369 & 0.382 & 0.385\\
  \midrule
  \multirow{3}{*}{H3N2} & ARAE & 0.171 & 0.210 & 0.219 & & 0.498 & 0.611 & 0.639 & & 0.255& 0.313& 0.326\\
  & GOAD & 0.152 & 0.178 & 0.190 & & 0.444 & 0.568 & 0.555 & & 0.226& 0.271& 0.283\\
  & ESAD & 0.279 & 0.290 & 0.331 & & 0.499 & 0.511 & 0.509 & & 0.358& 0.370 & 0.401\\
  & DeepSAD & 0.343 & 0.359 & 0.392 & & 0.536 & 0.540 & 0.565 & & 0.418 & 0.431 & 0.463\\
  \midrule
  \multirow{3}{*}{H5N1} & ARAE & 0.175 & 0.187 & 0.186 & & 0.488 & 0.505 & 0.512 & & 0.258& 0.273& 0.273\\
  & GOAD & 0.217 & 0.238 & 0.235 & & 0.538 & 0.566 & 0.569 & & 0.309& 0.335& 0.333\\
  & ESAD & 0.347 & 0.371 & 0.350 & & 0.568 & 0.538 & 0.482 & & 0.431& 0.439 & 0.406\\
  & DeepSAD & 0.418 & 0.453 & 0.471 & & 0.620 & 0.617 & 0.619 & & 0.499 & 0.522 & 0.535\\
  \bottomrule
\end{tabular}
\end{table}

\subsection*{Results}

We report the mean and standard deviation of the metrics that are used to evaluate our method in different experiments. The model is trained for 5 trials. Tables \ref{tab:auc}, \ref{tab:fscore}, \ref{tab:recall}, and \ref{tab:prec} show the performance of our method compared to Tempel \cite{yin2020tempel}, which is a recently proposed SOTA on the HA datasets. For each experiment, Area Under Curve (AUC), F1-Score, recall, and precision are reported as in other related works. To report Tempel's results, we use the original implementation that is publicly available on their reported link in their paper. Following the experiments on HA, we have also conducted similar experiments on the SARS-CoV-2 dataset for $T \in \{5,10\}$. Unlike the experiments on HA, for these experiments, $T$ presents time-series sequence length in months rather than years.

\textbf{Comparison of AUC, F1-Score, Recall, and Precision:} As it is shown, our method achieves competitive or significantly superior results in all standard criteria AUC, F1, recall, and precision averaged across all $T$ values on HA datasets. It is good to notice that our method is consistently better than Temple when the parameter $T$ is small, i.e. $T = 5$, on HA training sets for all different measures. This could be the effect of using an anomaly detection loss to train our model. Since DeepSAD loss attempts to find a shared representation space for the unmutated samples, it uses the given training set efficiently and, in the best scenario, extracts the most general features of them.
\begin{table}[!tbh]
\centering
\caption{AUC comparison on H1N1, H3N2 and H5N1 datasets for $T \in \{5,10,15\}$. Mean is the average of the results on T for each dataset and model. Our results are significantly better or competative with the SOTA method.}
\label{tab:auc}
\begin{tabular}{llcccc}
\hline
Dataset & Method &
\multicolumn{1}{c}{T=5} & \multicolumn{1}{c}{T=10} & \multicolumn{1}{c}{T=15} & \multicolumn{1}{c}{Mean} \\ \hline
\multirow{2}{*}{H1N1}
& Tempel\cite{yin2020tempel} & 0.8467 & 0.8537 & 0.8455 & 0.8486\\
& Ours   & \textbf{0.9713 $\pm$ 0.0002} & \textbf{0.9713 $\pm$0.00008} & \textbf{0.9715 $\pm$ 0.00015} & \textbf{0.9713}\\ \hline
\multirow{2}{*}{H3N2} & Tempel\cite{yin2020tempel} & 0.8989 & 0.8863 & 0.8884 & 0.8912\\
& Ours   & \textbf{0.9419 $\pm$ 0.00004} & \textbf{0.9419 $\pm$ 0.00004} & \textbf{0.9416 $\pm$ 0.0006} & \textbf{0.9418} \\ \hline
\multirow{2}{*}{H5N1} & Tempel\cite{yin2020tempel} & 0.9657 & 0.9696 & 0.9671 & 0.9674\\
& Ours & \textbf{0.9829 $\pm$ 0.00017} & \textbf{0.983 $\pm$ 0.00021} & \textbf{0.9819 $\pm$ 0.00026} & \textbf{0.9826}\\ \hline
\end{tabular}
\end{table}

\begin{table}[!tbh]
\centering
\caption{F1-Score of predictions on H1N1, H3N2 and H5N1 datasets for $T \in \{5,10,15\}$. Mean is the average of the results on T for each dataset and model.}
\label{tab:fscore}
\begin{tabular}{llcccc}
\hline
Dataset & Method &
\multicolumn{1}{c}{T=5} & \multicolumn{1}{c}{T=10} & \multicolumn{1}{c}{T=15} & \multicolumn{1}{c}{Mean} \\ \hline
\multirow{2}{*}{H1N1}
& Tempel\cite{yin2020tempel} & \textbf{0.8212}  & \textbf{0.821} & \textbf{0.8213} & \textbf{0.8211}\\
& Ours   & 0.8178 $\pm$ 0.0006 & 0.8176 $\pm$ 0.0004 & 0.81992 $\pm$ 0.0004 & 0.8184\\ \hline
\multirow{2}{*}{H3N2} & Tempel\cite{yin2020tempel} & 0.5957 & 0.5785 & 0.5913 & 0.5885 \\ & Ours   & \textbf{0.5966 $\pm$ 0.0009} & \textbf{0.5979 $\pm$ 0.003} & \textbf{0.60127 $\pm$ 0.001} & \textbf{0.5986} \\ \hline
\multirow{2}{*}{H5N1} & Tempel\cite{yin2020tempel} & 0.5167 & 0.5287 & 0.51722 & 0.5208\\ & Ours & \textbf{0.53083 $\pm$ 0.003} & \textbf{0.5387 $\pm$ 0.001} & \textbf{0.5287 $\pm$ 0.01} & \textbf{0.5327} \\ \hline
\end{tabular}
\end{table}

\begin{table}[!tbh]
\centering
\caption{Recall of predictions on H1N1, H3N2 and H5N1 datasets for $T \in \{5,10,15\}$. Mean is the average of the results on T for each dataset and model. Our results are competitive with SOTA on these datasets.}
\label{tab:recall}
\begin{tabular}{llcccc}
\hline
Dataset & Method &
\multicolumn{1}{c}{T=5} & \multicolumn{1}{c}{T=10} & \multicolumn{1}{c}{T=15} & \multicolumn{1}{c}{Mean} \\ \hline
\multirow{2}{*}{H1N1}
& Tempel\cite{yin2020tempel} & 0.8012 & 0.7997 & 0.8013 & 0.8007\\
& Ours   & \textbf{0.8033 $\pm$ 0.003} & \textbf{0.8066 $\pm$ 0.003} & \textbf{0.8071  $\pm$ 0.006} & \textbf{0.8056}\\ \hline
\multirow{2}{*}{H3N2} & Tempel\cite{yin2020tempel} & 0.5106 & 0.4793 & 0.4943 & 0.4947\\
& Ours   & \textbf{0.6107 $\pm$ 0.01} & \textbf{0.63305 $\pm$ 0.01} & \textbf{0.62768 $\pm$ 0.02} & \textbf{0.6238} \\ \hline
\multirow{2}{*}{H5N1} & Tempel\cite{yin2020tempel} & 0.3671 & 0.38181 & 0.36643 & 0.3717\\
& Ours & \textbf{0.4158 $\pm$ 0.01} & \textbf{0.4205 $\pm$ 0.009} & \textbf{0.42146 $\pm$ 0.01} & \textbf{0.4192}\\ \hline
\end{tabular}
\end{table}

\begin{table}[!tbh]
\centering
\caption{Precision of predictions on H1N1, H3N2 and H5N1 datasets for $T \in \{5,10,15\}$. Mean is the average of the results on T for each dataset and model.}
\label{tab:prec}
\begin{tabular}{llcccc}
\hline
Dataset & Method & \multicolumn{1}{c}{T=5} & \multicolumn{1}{c}{T=10} & \multicolumn{1}{c}{T=15} & \multicolumn{1}{c}{Mean} \\ \hline
\multirow{2}{*}{H1N1} & Tempel\cite{yin2020tempel} & \textbf{0.8423} & \textbf{0.8445} & \textbf{0.8389} & \textbf{0.8419}\\
& Ours   & 0.8330 $\pm$ 0.004 & 0.8292 $\pm$ 0.004 & 0.8333 $\pm$ 0.006 & 0.8318\\ \hline
\multirow{2}{*}{H3N2} & Tempel\cite{yin2020tempel} & \textbf{0.7147} & \textbf{0.7300} & \textbf{0.7358} & \textbf{0.7268}\\
& Ours   & 0.5845 $\pm$ 0.0004 & 0.5694 $\pm$ 0.01 & 0.5786 $\pm$ 0.01 & 0.5775\\ \hline
\multirow{2}{*}{H5N1} & Tempel\cite{yin2020tempel} & \textbf{0.8721} & \textbf{0.8598} & \textbf{0.8791} & \textbf{0.7803}\\
& Ours   & 0.7530 $\pm$ 0.02 & 0.7606 $\pm$ 0.02 & 0.7176 $\pm$ 0.02 & 0.7437 \\ \hline
\end{tabular}
\end{table}

\textbf{Stopping Criterion of the Training Process: }As mentioned before, we use validation data to find the best point to cut the training process. Figure S3 shows the F1-score curves of the validation and test tests. As is depicted, the validation curves almost always follow their corresponding test curves, which shows their usability in determining the points to stop the training process. Note that the smaller the value of parameter $T$, the more the complexity of finding a good representation space. Therefore, a larger number of training samples is needed to approximately make the validation and test distributions similar, and consequently, some validation curves do not completely follow the test curves.

\textbf{Comparison of ROC curves: } Figures S4 and S5 show the ROC curves of our methods compared to Tempel. As it is obvious, our method always has a lesser value of false-positive rate for high true positive rate values for all $T$ values on all HA test datasets. This can be justified by the margin that exists between the normal and abnormal distributions. The HSC loss attempts to make this margin as large as possible, but classification-based approaches such as Tempel only try to minimize the cross-entropy loss that does not consider maximizing the margin explicitly. This helps our model to be more robust against noises that exist in our training datasets, which are typical considering the difficulties of the dataset making process. Moreover, for SARS-CoV-2 dataset, our method, in setups with lower false-positive rates, shows a higher true-positive rate compared to Tempel.

\textbf{Ablation study results: } We have also conducted further experiments using a different model called transformers for predicting the mutations. The results for these experiments on HA are shown in table \ref{tab:transformer}. Details about the mechanism of the transformers are provided in the ablation section of methods.

\begin{table}[!tbh]
\centering

\caption{Recall, F1-score and AUC on H1N1, H3N2 and H5N1 datasets for $T \in \{5,10,15\}$. Results are for The Transformer Network and Tempel}
\label{tab:transformer}
\begin{tabular}{@{}lllllllllllll@{}}
  \toprule
  Dataset & Method &\multicolumn{3}{c}{Recall} & & \multicolumn{3}{c}{F1-score} & &\multicolumn{3}{c}{AUC} \\
  \midrule
  & & 5 & 10 & 15 & & 5 & 10 & 15 & & 5 & 10 & 15 \\
  \cmidrule{3-5} \cmidrule{7-9} \cmidrule{11-13}
  \multirow{2}{*}{H1N1} & Tempel & 0.80128 & 0.79972 & 0.80137 & & 0.82128 & 0.82102 & \textbf{0.82138} & & 0.84677 & 0.85376 & 0.84551\\
     & Transformer & \textbf{0.8024} & \textbf{0.80137} & \textbf{0.80432} & & \textbf{0.82151} & \textbf{0.8218} & 0.81867 & & \textbf{0.9717} & \textbf{0.9711} & \textbf{0.96935}\\ \hline
  \midrule
  \multirow{2}{*}{H3N2} & Tempel & 0.51069 & 0.479381 & 0.49432 & & 0.59571 & 0.57859 & \textbf{0.59137} & & 0.89896 & 0.88632 & 0.88847\\
     & Transformer & \textbf{0.63373} & \textbf{0.632757} & \textbf{0.5949} & & \textbf{0.6196} & \textbf{0.61002} & 0.5902 & & \textbf{0.9445} & \textbf{0.9354} & \textbf{0.91917}\\ \hline
  \midrule
  \multirow{2}{*}{H5N1} & Tempel & 0.3671 & 0.38181 & 0.36643 & & 0.5167 & \textbf{0.52878} & \textbf{0.51722} & & 0.96572 & 0.9696 & 0.96715\\
     & Transformer & \textbf{0.38083} & \textbf{0.43426} & \textbf{0.42167} & & \textbf{0.5174} & 0.48174 & 0.49658 & & \textbf{0.98874} & \textbf{0.97638} & \textbf{0.97603}\\ \hline
  \bottomrule
\end{tabular}
\end{table}

\subsubsection* {Mutation prediction results in the real world dataset assessment}
After the evaluation of possible amino acid alterations, mutations that exhibit top recall and precision rates for T=5, T=10, and T=15 are listed and assessed based on random samples of the 2016 reported influenza sequences and the 2016-2017 season vaccine strain. Table S1 represents the model's results for different possible amino acid alteration locations and their presence in the random sampling data from the 2016 reported stains. It should be noted that based on the model's results, every reported alteration, as an amino acid number is associated with its previous and next amino acid. The proposed alterations by the algorithm in sampling data are evaluated separately and regardless of the algorithm ROC curve for the mutations. Furthermore, the proposed alteration positions that do not depict any mutations in figures S1 and S2 or table S1 in the sampling aligned data from the 2016 year do not express that there is no mutation in all of the years of the data set for that particular position. The mutations in these particular locations do not appear in the current random sampling due to the low stability or frequency of the mutation.

\textbf{H1N1 influenza A: }
By the evaluation of the antigenic sites of influenza H1N1 amino acid sequence, the proposed model highlights some important alterations in amino acid numbers 168, 170, 205 and 253 in three important sites Ca, Sa and Sb. Meanwhile, there are too many missed important amino acid alterations in Sa, Sb, and specially in Ca. More details are provided in figure S1. In the comparison of our current anomaly detection method with the suggested method by Tempel, there are three differences in alteration positions. These differences refer to the amino acid numbers 165 and 174 by Tempel and 166 in our model. The residues 165 and 166 did not represent any important alterations while 174 is located in Sa.\\

\textbf{H3N2 influenza A: }
By the evaluation of the antigenic sites of influenza H3N2 amino acid sequence, the model proposed some important alterations in amino acid number 192 in antigenic Site B and amino acid 226 in the Receptor binding site (RBS). Some alterations in other locations are illustrated in figure S2. It should be mentioned that some important alterations in antigenic sites Lower HA, site A, and Site B are missed. In the comparison of our model and Tempel, each model proposed two positions that the other model did not detect. The residues 88 and 260 are suggested by the Tempel model and not by our current model. The residue 88 represents variable positions in the sequence but not epitope. While amino acid number 260 is an epitope location and missed by our model. Our model suggests alteration in the amino acids numbers 177 and 193 more than the Tempel model. The amino acid 177 does not reflect any epitope or high prevalence variation, while the 193 is a true mutation in the site B.\\

\textbf{H5N1 influenza A: }
The influenza H5N1 amino acid sequence was not evaluated in the reported sequences from the 2016 human host due to the lack of reports in that particular year. For the evaluation of the model and multiple sequence alignment for more important alteration positions in table S1, we used the avian sequences of the H5N1 during 2016 due to the nature of the virus. The amino acids numbers 82, 116, 152, 166, 172, 179, 185, 205, and 242 seem to be important for further studies and antigenic evaluation. Moreover, by comparing our current model and the Tempel model, there are two different mutation locations in each (118 and 130 for Tempel in comparison with 172 and 179). The conducted study reveals two important mutation positions in SARS-CoV-2. The data was collected on 8 June 2021. The proposed mutation positions are the 476 and 500 amino acid locations. The amino acid number 500 represents the Alpha variant-specific mutation (N501Y) on the date of the dataset preparation. Another proposed mutation position in 476 did not represent any particular association with any variants except the Omicron variant (S477N). It has to be noted that at the time of the database preparation, there was no clue about the Omicron variant, and this mutation in the 476 positions did not reflect any particular result.\\



\section*{Discussion}
Every year, hundreds of thousands of deaths are reported from the influenza disease. Despite many efforts to develop new treatments and vaccines, due to the high genetic diversity of the influenza viruses (IVs), complete success has not been achieved yet. This extended genetic diversity is due to the RNA genome of these viruses. Between the proteins encoded by the IVs genome, hemagglutinin (HA) is among the most important proteins that play key roles in the infectivity and propagation of the virus \cite{wu2006mutation}. This glycoprotein plays a paramount role in binding to the Sialic acid, the ligand of the virus at the surface of the host cell, as well as fusion into the cell. In fact, the pathogenicity of IVs depends on efficient cleavage of hemagglutinin precursor to HA1 and HA2 proteins, which the former is responsible for receptor-binding activity and the latter anchors the HA1 and is also responsible for pH-dependent fusion \cite{shirvani2020contributions}. This reason, along with the increased resistance to currently available classes of drugs, which inhibit M2 ion channels and neuraminidase, has made inhibition of hemagglutinin one of the attractive goals for the development of anti-influenza drugs in recent years \cite{shen2013novel}. In addition to the medications, the immune system uses HA as a target in response to infection. Once the infection is established, the adaptive immune system triggers a strong response against the virus, in which neutralizing antibodies are produced against the virus. HA is the major target of these neutralizing antibodies \cite{knipe2013fields}. Since these antibodies are strain specific \cite{krammer2013influenza}, mutations in the HA epitopes that are targeted by these antibodies may change the antigenicity of the virus and lead to the emergence of antibody- and vaccine-escape strains \cite{knipe2013fields, ning2019antigenic}. After entrance of viruses to the body and production of neutralizing antibodies by the immune cells, antibody-escape viruses evade the host immune system, a phenomenon called selection of “antibody escape” variants \cite{knipe2013fields}. \\
Influenza virus is highly prone to antigenic changes. These changes are often caused by two main mechanisms called the antigen shift and drift. If a cell is infected with two different genotypes of the IVs, a new strain may develop due to the placement of different parts of the two strains into new viral particles. This phenomenon, which can lead to generation of new pandemic strains, is called the antigen shift. Influenza viruses are highly prone to point mutations due to the lack of proof-reading ability of their RNA genome. The accumulation of these point mutations is called the antigen drift. Occurrence of this phenomenon in the gene encoding HA causes alterations in the structure and function of this protein \cite{de2018molecular}. As a result, the immune system is no longer able to detect the virus \cite{webster1982molecular}. Therefore, this jeopardizes us at risk of future pandemics to which our bodies have no resistance \cite{wu2005timing}. \\
Since HA has a critical role in the replication cycle of IVs, occurrence of these changes have greatly contributed to the evolution of these viruses \cite{doud2018single}. A series of mutations with few effects on antigenicity of the IVs finally lead to intense antigenic drift, a phenomenon called “cluster transition,” which is the characteristic of evolution of IVs \cite{lyons2018mutation}. Studies have shown that changes in the amino acid sequence of HA1 can tremendously alter the antigenicity of these viruses \cite{webster1982molecular}. Thereby tracking these changes is essential to predict the future behaviors of the virus. Thyagarajan and Bloom mutagenized HA gene of wild type IVs to create codon-mutant libraries of HA gene and then used these libraries to make a pool of mutant by reverse genetics. Using the Illimina deep mutational scanning, they found that there are more than 10,000 different probabilities for mutation in this protein. They have also cultured these mutants and investigated the mutated viruses. They have concluded that mutations mostly occur in the regions of the HA protein that is recognized by antibodies. Otherwise, the receptor-binding domain is less frequently subject to mutations and it is the possible reason that why HA still targets the Sialic acid in spite of extensive mutations in this protein \cite{thyagarajan2014inherent}.\\
Determining the new antigenic variants of the IVs is critical in order to develop efficient flu vaccines. The WHO collaborates with a number of laboratories around the world to identify circulating influenza viruses in the human population (\cite{morris2018predictive}. Traditionally, the hemagglutination inhibition (HI) test has been used to evaluate antigenic variants of the virus, which is a time-consuming and hand-operated method \cite{liao2008bioinformatics}. However, advances in computer science and bioinformatics have led to the invention and development of faster and more accurate methods. By using scoring and regression methods on the sequences collected from the H3N2 flu virus between 1971 and 2002, Liao et al. proposed a method for predicting variants of the virus. According to their proposed method, influenza virus variants between 1999 and 2004 were predictable with an agreement rate of 91.67$\%$. They also identified 20 amino acid positions whose changes significantly contribute to the development of new variants \cite{liao2008bioinformatics}. Yang et al. developed a learning sparse algorithm called AntigenCO, which uses the HA1 protein of the H3N2 virus to identify the superior determinant properties of the antigenic profiles in serological data. In addition to single mutations, their methods also used multiple simultaneous mutations or co-evolved sites in order to predict the antigenicity. The prediction accuracy of the method studied in their work was 90.05$\%$ \cite{yang2014sequence}. Łuksza et al. developed a model based on mapping between the HA sequences and viral fitness. According to them, mutation at the epitope region is considered a positive fitness effect as it can induce cross-immunity between the flu strains. Conversely, the incidence of mutation outside of the epitope site is regarded as a fitness cost. Using this system, viral clades that are involved in future epidemics can be predicted \cite{luksza2014predictive}. Yin et al. developed a computational method to predict suitable strains for vaccination by generating time-series training samples with splittings and embeddings. Their method uses the Recurrent Neural Network (RNN), which helps in performing sequence to sequence prediction using H3N2 flu strains identified between 2006 and 2017. The suggested strains by their method had 98$\%$ similarity to the recommended vaccine strains \cite{yin2020time}.
In the current work, we study an anomaly detection-based approach in predicting the Influenza mutations. The study results suggest some important alteration positions in the HA of the influenza H1N1, H3N2, and H5N1. This multidisciplinary study shows promising and reliable results in the mutation prediction in influenza virus in comparison with mentioned previous studies. Based on the nature of the anomaly detection, this approach seems to be appropriate for mutation prediction analysis in viral genomes for further studies. By considering the complete HA gene and the ROC curves, our current model represents great performance. However, in the assessment of important antigenic sites, more optimization would be critical for further studies. Furthermore, one of the major limitations in the current study was the limited available sequences for different years. This limitation reflects a minor problem in our current research by considering the parameter T. Meanwhile, great efforts in the time of the SARS-CoV-2 pandemic and availability of a great amount of the primary sequence data could be promising for further studies. By considering the advancement of high throughput sequencing techniques, this limitation in sequences can be solved in future. Another limitation in our current study was the evaluation of all lineages and clades of any H type influenza. Different clades or linages in H1N1 have unique evolutions and need more focus in future studies in light of the improving primary sequence limitation. In addition, another limitation in the analysis of the H5N1 mutations, which needs to be mentioned, is the lack of the H5N1 reports in human hosts during 2016.\\
Recently, this RNN model is used to predict the antigenic changes of the Severe respiratory syndrome coronavirus 2 (SARS-COV-2) by Sawmya et al\cite{sawmya2021analyzing}. They used a pipeline method to predict mutations in some regions of the SARS-COV-2 gene. At first, genes are classified according to the country, and their phylogenetic tree is obtained. In the next step, using algorithmic methods and methods used in machine learning, positions of genes that are distinct in terms of characteristics are obtained. Finally, the occurrence of mutations in these sites is predicted by using the CNN-RNN network. Considering the SARS-CoV-2, the current study’s data is just preliminary and just a perspective for further studies. This preliminary data represents low productivity in all other potential positions for mutations in the SARS-CoV-2 S gene. But highlights an important position for mutation in amino acid number 500 (represents N501Y mutation in Alpha variant). 

\section*{Methods}
\subsection*{Data availability}
The data that support the findings of this study are available from Tempel\cite{yin2020tempel}. The data is hosted at \url{https://drive.google.com/drive/folders/1-pJGBsVfIqCEizetTQe43OQJvkmhocdW}.
\subsection*{The influenza virus HA and SARS-CoV-2 RBD database for amino acid sequences}
The influenza virus sequences are obtained from the NCBI influenza database \url{http://www.ncbi.nlm.nih.gov/genomes/FLU/FLU.html}. The amino acid sequences were obtained without the limitation in time for the report. The amino acid sequences for the human hosts in the north hemisphere referring to the H1 of the H1N1 strains were downloaded. Only the full-length sequences for the HA region were used for the Multiple Sequence Alignment (MSA). The MSA is performed by the MAFFT algorithm.
For the experiments on SARS-CoV-2, we used the COVID-19 data portal provided by the European COVID-19 Data Platform and EMBL’s European Bioinformatics Institute (EMBL-EBI)\cite{coviddata}. 
\subsection*{Proposed method}
The mutation prediction problem can be formulated as the time series anomaly detection. For getting the time series data out of the raw amino-acid sequence, a preprocessing method is employed. Firstly, the ambiguous amino acids are replaced with one of the common 20 amino acids. Then, the amino acid sequences are clustered, and the primary time series sequence is obtained from them. Lastly, the amino acids are represented in some feature space using the prot2vec\cite{asgari2015continuous} method. In order to formulate the problem as the anomaly detection, we  selected temporal attention-based RNN \cite{qin2017dual} as the backbone that is trained with a DeepSAD loss function. This lets us not only preserve the sequential aspect of the input, but also benefit from the anomaly detection algorithms. Temporal attention-based RNN has shown significantly better performance compared to LSTM networks in preserving long-term dependencies. Besides, owing to the use of attention mechanism, it only focuses on the important features of its input, which is helpful in complex problems such as the  mutation prediction tasks. In addition, DeepSAD loss function helps our model to not only focus on the important parts of the input sequence but also encapsulate them in a compact representation space. This is exactly what we look for in the mutation prediction problem. Our codes are made publicly available at \url{https://github.com/rohban-lab/Tempel-HSC-} for the sake of reproducibility.
\subsection*{Preprocessing}
The preprocessing, similar to Yin et al. \cite{yin2020tempel}, includes three major steps: 1. replacing obscure amino acids; 2. acquiring time series sequences; 3. representing the amino acids in a feature space. Proteins consist of 20 common amino acids. Due to errors in reading the protein sequence in the stated datasets, there might be some ambiguous amino acids or letters, such as 'B', 'Z', 'J', and 'X', in the sites, that are needed to be replaced with a probable amino acid. In this step, 'Z' will be randomly replaced with one of 'D' or 'N' amino acids, 'Z' with one of 'E' or 'Q', 'J' with one of 'I' or 'L', and 'X' with all amino acids. After replacement, for obtaining the time series sequences, samples in each year are divided into $k$ clusters using the $k$-means algorithm. Then, by a similarity metric such as the Euclidean distance, for each cluster in a year, the closest cluster in the next year is selected. By finding the closest cluster pairs in two consecutive years, multiple sequences of clusters are formed. For example, assume that in the year $t$, the cluster $M$ is closest to the cluster $N$ in the year $t+1$, and the cluster $O$ in the year $t+2$ is closest to the cluster  $N$. So $M, N, O, \dots$ will make a cluster sequence. From each cluster in a cluster sequence, a protein sample is randomly selected, and hence from this sequence, a time series sequence of data is created. In feature space representation, each amino acid is converted to a vector using ProtVec\cite{asgari2015continuous}. The representation is calculated by the following equation:
\begin{ceqn}
\begin{align}
x^t_i = \frac{\sum\limits_{j = -N}^{N} v_{i+j}^t}{2N+1}
\label{eq:rep}
\end{align}
\end{ceqn}
where $v_i^k$ is the vector output of ProtVec for the $i$-th amino acid in a sample gathered in the year $t$. $N$ shows the number of amino acid neighbors. Here, like Yin et al., representations are evaluated in 3-grams, so $N$ is set to 1. After these steps, specific site positions, which are also known as epitopes, are only selected in each protein sequence, and the RNN model is only trained and tested on these sites.
\subsubsection*{Temporal Attention}
Given a specific time series sequence $(x_1, x_2,\dots,x_t)$, a temporal attention mechanism is used for encoding the sequence. Architectures such as LSTM and GRU have a performance decay when the input length increases. By applying temporal attention mechanism to the mentioned architectures, we are able to consider each element of the sequence in the final encoding of the sequence and overcome this issue.
In this mechanism, in order to calculate attention weights $w_{ti}$,  first, an attention function $f$ takes the hidden state $h_{i}$ for the time $i$ and the cell state $s_{t-1}$ for the time $t-1$ from the LSTM cell and outputs a scalar value $e_{ti}$:
\begin{ceqn}
\begin{align}
e_{ti} = f(s_{t-1}, h_i).
\label{eq:eti}
\end{align}
\end{ceqn}

Then using a softmax function, the weight $w_{ti}$ is calculated and applied on the hidden states as bellow:
\begin{ceqn}
\begin{align}
w_{ti} = \frac{\exp(e_{ti})}{\sum\limits_{j=1}^{t-1} \exp(e_{tj})}\\
c_t = \sum\limits_{j=1}^{t-1} w_{tj}h_j,
\label{eq:eti}
\end{align}
\end{ceqn}
where $c_t$ is the context vector for the time $t$. In the last step, using the following equation, the encoded vector $\hat{h}_t$ at time $t$ is calculated:
\begin{ceqn}
\begin{align}
\hat{h}_t = tanh(W^{}_{\hat{h}}[c^{}_{t} ; h^{}_{t}] + b^{}_{\hat{h}}),
\label{eq:eti}
\end{align}
\end{ceqn}
where $W^{}_{\hat{h}}$ and $b^{}_{\hat{h}}$ are learned through the training process. In the next step, the DeepSAD loss discussed in the next section, is applied to the encoded vector $\hat{h}_{T}$, where $T$ is the length of the time series sequence.

\subsubsection*{DeepSAD Loss}
As mentioned earlier, we use DeepSAD loss function\cite{ruff2019deep} on temporal attention RNN to capture normal features. DeepSAD loss function is defined in Eq. \ref{deep_sad_loss}. 

\addtolength\abovedisplayskip{0.8\baselineskip}
\addtolength\belowdisplayskip{1.2\baselineskip}

\begin{ceqn}
\begin{align}
\label{deep_sad_loss}
\dfrac{1}{m + n} \sum_{i=1}^n   \Vert\phi(\hat{h}_{i}; \theta) - c\Vert^{2} + \dfrac{\eta}{m + n} \sum_{j=n+1}^{m+n}  (\Vert\phi(\hat{h}_{j}; \theta) - c\Vert^{2})^{-1} + \dfrac{\lambda}{2} \sum_{l=1}^L  \Vert\theta^{l}\Vert^{2}_{F}
\end{align}
\end{ceqn}

Here, $\phi(.; \theta)$ is a neural network, our temporal attention-based RNN, and $\theta$ denotes the parameters of $\phi$. Also, $c$ is a hyper-parameter that is set before the training process. By minimizing the loss function, normal samples are centralized in a minimum volume hyper-sphere with a center $c$, while  abnormal samples are forced to be out of it. Also, $\eta$ is a hyper-parameter that can assign extra weight to the abnormal samples to mitigate the imbalanced training samples issue. Finally, $n$ and $m$ denote the number of normal and abnormal training samples,  respectively. 

As mentioned in Ruff et al. \cite{ruff2020rethinking} use of the radial basis function in the DeepSAD loss produces better results. Therefore, the Eq. \ref{HSC} that is called the hyper-sphere classifier (HSC) loss function is used in all our experiments:
\begin{ceqn}
\begin{align}
\label{HSC}
\dfrac{1}{m + n} (\sum_{i=1}^{m+n} y^{}_{i} \Vert{\phi(\hat{h}_{i}; \theta)}\Vert^2 - (1 - y^{}_{i}) \log(1 - \exp(-(\sqrt{\Vert{\phi(\hat{h}_{i}; \theta)}\Vert^2 + 1} - 1))))
\end{align}
\end{ceqn}

HSC makes normal data to be mapped near a defined center. Since all data in our dataset has a label, the dataset is assumed to be a set of pairs $(x^{}_{i}, y^{}_{i})$. Here, $m + n$ is the number of training samples, and  $y \in$ \{0, 1\}. Finally, $y=1$ shows normal data (in our case, an unmutated virus), and $y=0$ shows a mutated sample.

\subsubsection*{Ablation}
As described above, RNNs are one of the mechanisms that can produce satisfactory results in the case of sequence transduction. However, with all their efficiencies, they have the problem of capturing long-term dependencies, which is mainly because of the fact that input sequences are passed at each step in a chain, and if the chain becomes longer, it will be more probable that the information gets lost. Unlike RNNs, Transformers can be very efficient in retaining the long-term dependencies.  Transformers are a kind of Neural Network architecture that has become popular due to their performance in various fields. They were introduced to perform sequence transduction in which there are dependencies between input elements. Its architecture consists of an encoder and a decoder. The encoder has some layers that together generate encodings that summarize each element in relation to all other elements in the sequence, through the self-attention mechanism. Each layer consists of a self-attention mechanism and a Feed-Forward Neural Network.

When an input sequence is fed into the encoder, each input part flows through each of the two layers of the encoder. They first go through a self-attention mechanism, which learns three weight matrices; the query weights $W_{Q}$, the key weights $W_{K}$, and the value weights $W_{V}$. For each input element $i$, the input $x_{i}$ is multiplied with each of the three vectors. This will produce a query vector $q_{i} = x_{i}W_{Q}$, a key vector $k_{i} = x_{i}W_{K}$, and a value vector $v_{i} = x_{i}W_{V}$. The attention weight $a_{ij}$ from the input element $i$ to the input element $j$ can be obtained by applying the dot product between $q_{i}$ and $k_{j}$. Then, the attention weights are divided by $d_{k}$, which is the dimension of the key vectors. This leads to having more stable gradients, which is an advantage over RNNs. Then, we pass the results through a softmax function in order to normalize the weights. Next, we multiply each value vector by the softmax score and at the end, we sum up these weighted value vectors as follows:
\begin{ceqn}
\begin{align}
\label{self-attention}
Attention(q_{i}, k_{i}, v_{i}) = Softmax \left({\frac {q_{i} k^{T}_{i}}{\sqrt{d_{k}}}}\right) v_{i}.
\end{align}
\end{ceqn}
The resulting vector would then be sent along with the Feed Forward Neural Network. The decoder functions in a similar way. The decoder module contains some layers that take all the encodings and use them to generate an output sequence. Each decoder layer has both components of an encoder layer but it also has an attention layer between the two components. This helps the decoder focus on relevant parts of the input sequence.

\section*{Conclusion and Future Work}

In this research, we have shown the effectiveness of anomaly detection approaches in predicting Influenza mutations by conducting extensive experiments. Due to the extreme imbalance   between the training samples in such problems, anomaly detectors can extract useful information more efficiently and find shared, more robust features for the unmutated samples. Also, the results when the parameter $T$ is small advocate the mentioned merits of our approach. For future works, we want to extend our experiments and provide some results on SARS-CoV-2 (severe acute respiratory syndrome coronavirus 2) mutation prediction, which can be very helpful in understanding its behavior for further applications.\\

\clearpage

\bibliography{sample}

\end{document}


\section*{Supplementary Materials}
    \subsection*{Figures}
        \begin{figure*}[!htb]
          \centering
          \begin{subfigure}[b]{0.55\textwidth}
             \centering
             \includegraphics[width=1\linewidth, scale=0.1]{./Images/H1N1 epitops 1.png}
             \caption*{}
          \end{subfigure}
          \begin{subfigure}[b]{0.55\textwidth}
             \centering
             \includegraphics[width=1\linewidth, scale=0.1]{./Images/H1N1 epitops 2.png}
             \caption*{}
          \end{subfigure}
          \begin{subfigure}[b]{0.55\textwidth}
             \centering
             \includegraphics[width=1\linewidth, scale=0.1]{./Images/H1N1 epitops 3.png}
             \caption*{}
          \end{subfigure}
          \caption{The antigenic structure and the proposed mutations by the current study for H1N1. Symbols represents the proposed alteration by the model in right location (\checkmark), not detected important alteration in antigenic site ($\boxtimes$), and proposed alteration by model that does not reflect any important or high prevalence alteration (\square).}
          \label{fig:mut_h1n1}
\end{figure*}
        
        \begin{figure*}[!htb]
          \centering
          \begin{subfigure}[b]{0.475\textwidth}
             \centering
             \includegraphics[width=1\linewidth, scale=0.1]{./Images/H3N2 epitop 1.png}
             \caption*{}
          \end{subfigure}
          \begin{subfigure}[b]{0.475\textwidth}
             \centering
             \includegraphics[width=1\linewidth, scale=0.1]{./Images/H3N2 epitop 2.png}
             \caption*{}
          \end{subfigure}
          \begin{subfigure}[b]{0.475\textwidth}
             \centering
             \includegraphics[width=1\linewidth, scale=0.1]{./Images/H3N2 epitop 3.png}
             \caption*{}
          \end{subfigure}
          \begin{subfigure}[b]{0.475\textwidth}
             \centering
             \includegraphics[width=1\linewidth, scale=0.1]{./Images/H3N2 epitop 4.png}
             \caption*{}
          \end{subfigure}
          \caption{The antigenic structure and the proposed mutations by the current study for H3N2. Alignment numbers are set based on the HA gene without the first 16 amino acids of the signal peptide in the figure for illustration of the antigenic locations. While in the assessment by the algorithm, we evaluate this 16 amino acids and represented numbers for locations in table S1.\\
        Symbols represent the proposed alteration by the model in right location (\checkmark), not detected important alteration in antigenic site ($\boxtimes$), and proposed alteration by the model that does not reflect any important or high prevalence alteration (\square).
        }
          \label{fig:mut_h3n2}
        \end{figure*}
        
        \begin{figure*}[!htb]
  \centering
  \includegraphics[trim=3cm 2cm 3cm 2cm,clip, width=1\linewidth]{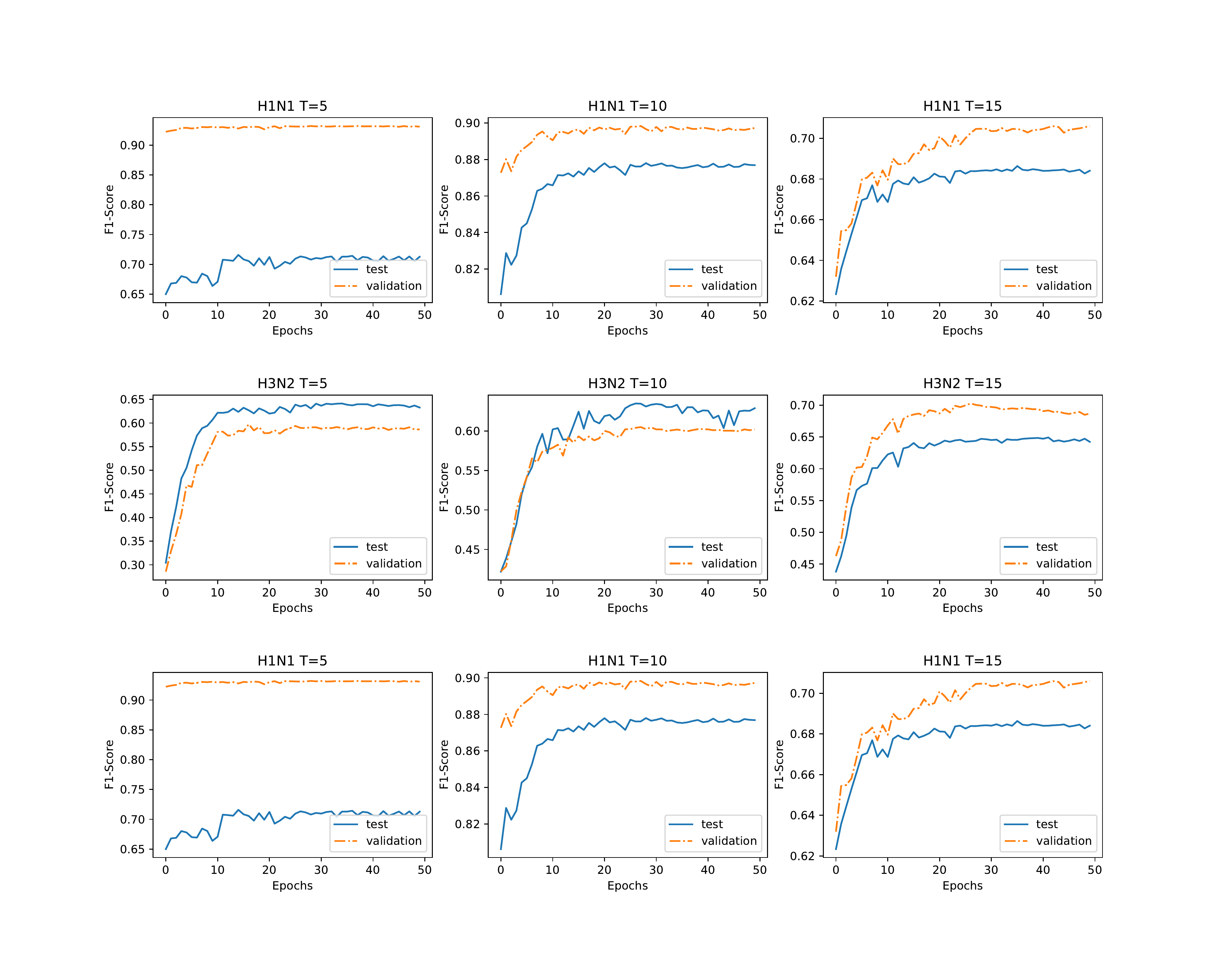}
  \caption{F1-Score of each experiment over epochs. The validation F1-Score in every epoch is achieved by adjusting the threshold in such a way that maximizes this score on the validation data. Later in the same epoch, this threshold is applied to the test dataset to get the test F1-Score.}
  \label{fig:stab}
\end{figure*}

        \begin{figure*}[!htb]
  \centering
  \includegraphics[trim=3cm 2cm 3cm 3cm,clip, width=1\linewidth]{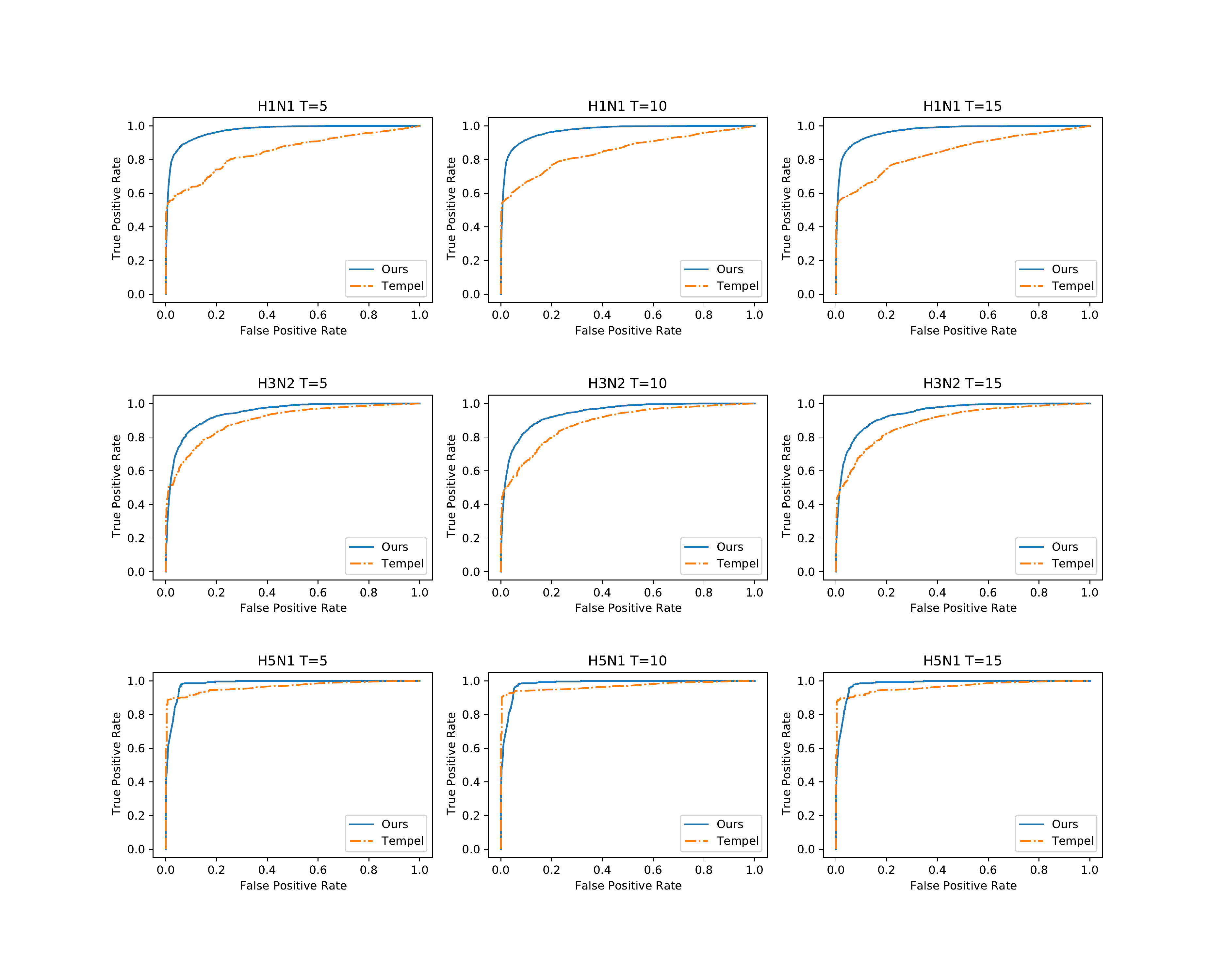}
  \caption{The ROC curves of our model and Tempel for each experiment on the HA dataset.}
  \vspace{180in}
  \label{fig:roc}
\end{figure*}

        \begin{figure*}[!htb]
  \centering
  \includegraphics[ width=1\linewidth]{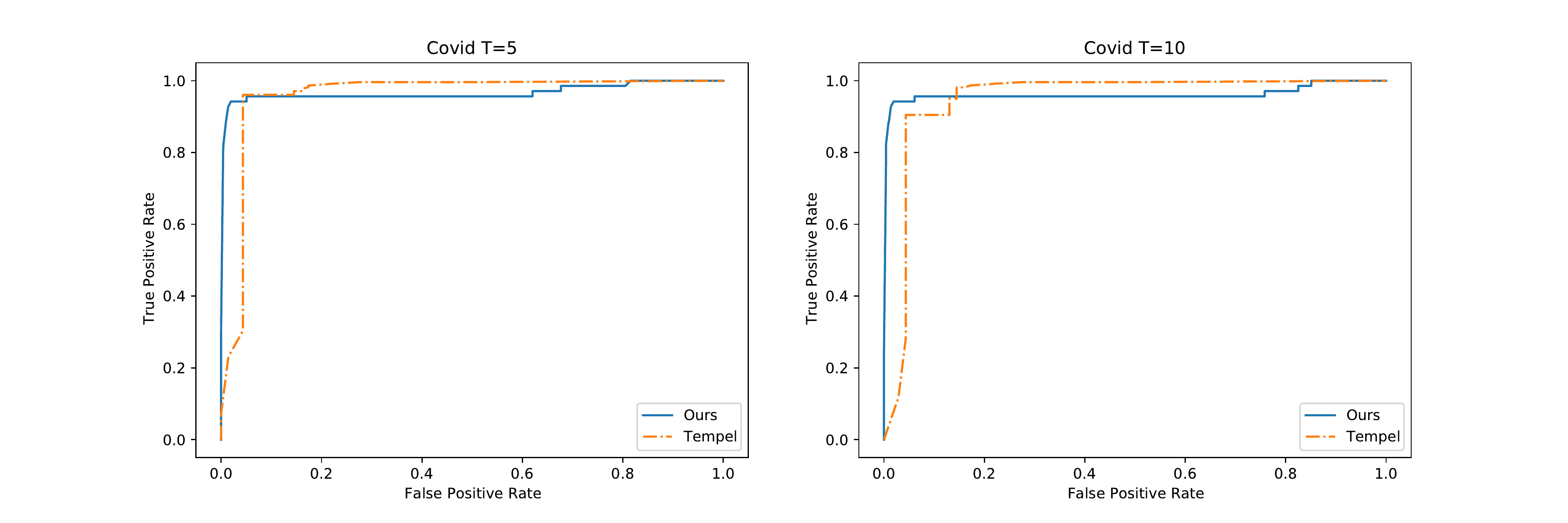}
  \caption{The ROC curves of our model and Tempel for each experiment on the SARS-CoV-2 dataset.}
  \vspace{180in}
  \label{fig:roc_cov}
\end{figure*}
        \FloatBarrier
    \subsection*{Tables}
        \begin{figure}[!htb]
\begingroup
  \footnotesize 
  \setlength\LTleft{15pt}%
  \setlength\LTright{15pt}%
  \centering 
  \begin{longtable}{@{\extracolsep{\fill}}|c |c| c| c| c| c| c| c| c@{}|}
     \caption{The model's results for different possible amino acid alteration locations.\label{tab:other_methods}}\\
    \hline\hline 
  Virus type & Mutation &\multicolumn{3}{c}{Current study} & & \multicolumn{3}{c}{Tempel} \\
  \midrule
  & & Mutation & Mutation & Mutation^{*} & & Mutation & Mutation & Mutation^{*}\\
  & & Recall & Precision &  & & Recall & Precision &  \\
  \cmidrule{3-5} \cmidrule{7-9}
  \multirow{3}{*}{H1N1(T5, T10, T15)} & 69 & 0.92 & 0.97 & Y & & 0.95 & 0.97 & Y\\
  & 70 & 0.93 & 0.95 & N & & 0.9 & 0.95 & N\\
  & 85 & 0.89 & 0.95 & Y & & 0.89 & 0.97 & Y\\
  & 95 & 0.97 & 1 & N & & 0.97 & 1 & N\\
  & 132 & 0.93 & 0.89 & N & & 0.95 & 0.9 & N\\
  & 139 & 0.89 & 0.95 & N & & 0.95 & 1 & N\\
  & 140 & 0.91 & 0.91 & N & & 0.89 & 0.91 & N\\
  & 142 & 0.92 & 1 & N & & 0.92 & 1 & N\\
  & 152 & 0.91 & 1 & Y & & 0.93 & 1 & N\\
  & 165 & - & - & - & & 0.95 & 0.97 & Y\\
  & 166 & 0.84 & 0.97 & N & & - & - & -\\
  & 168 & 0.97 & 0.95 & N & & 0.9 & 0.95 & N\\
  & 170 & 0.88 & 0.97 & Y & & 0.88 & 0.97 & Y\\
  & 174 & - & - & - & & 0.9 & 0.92 & Y\\
  & 205 & 0.95 & 0.93 & Y & & 0.93 & 0.93 & Y\\
  & 226 & 0.9 & 0.97 & Y & & 0.89 & 0.95 & Y\\
  & 253 & 0.93 & 0.97 & Y & & 0.93 & 0.97 & Y\\
  & 276 & 0.86 & 0.97 & Y & & 0.9 & 0.97 & Y\\
  \midrule
    \multirow{3}{*}{H3N2(T5, T10, T15)} & 44 & 0.75 & 0.5 & N & & 1 & 1 & N\\
  & 88 & - & - & - & & 0.84 & 1 & N\\
  & 177 & 1 & 0.83 & N & & - & - & -\\
  & 193 & 1 & 0.85 & Y & & - & - & -\\
  & 197 & 1 & 0.85 & Y & & 0.5 & 1 & Y\\
  & 198 & 0.58 & 0.31 & Y & & 0.58 & 0.77 & Y\\
  & 201 & 0.5 & 0.28 & N & & 0.25 & 0.5 & N\\
  & 203 & 0.66 & 0.77 & N & & 0.41 & 1 & N\\
  & 207 & 1 & 0.8 & N & & 0.5 & 1 & N\\
  & 208 & 0.87 & 0.85 & N & & 0.57 & 1 & N\\
  & 209 & 0.75 & 0.6 & N & & 0.5 & 1 & N\\
  & 212 & 0.45 & 0.62 & Y & & 0.36 & 1 & Y\\
  & 213 & 0.46 & 0.8 & N & & 0.28 & 1 & N\\
  & 214 & 0.57 & 0.75 & N & & 0.3 & 1 & N\\
  & 215 & 0.8 & 0.5 & N & & 0.4 & 0.67 & N\\
  & 216 & 0.69 & 0.5 & N & & 0.53 & 1 & N\\
  & 217 & 0.9 & 0.76 & N & & 0.81 & 1 & N\\
  & 218 & 0.95 & 0.87 & N & & 0.45 & 1 & N\\
  \midrule
   
      \multirow{3}{*}{H5N1(T5, T10, T15)} & 82 & 1 & 1 & Y & & 1 & 1 & Y\\
  & 91 & 1 & 1 & N & & 1 & 1 & N\\
  & 116 & 1 & 1 & Y & & 1 & 1 & Y\\
  & 118 & - & - & - & & 1 & 1 & N\\
  & 130 & - & - & - & & 1 & 1 & Y\\
  & 152 & 1 & 1 & Y & & 1 & 1 & Y\\
  & 162 & 1 & 1 & N & & 1 & 1 & N\\
  & 163 & 1 & 1 & N & & 1 & 1 & N\\
  & 165 & 1 & 0.5 & N & & - & - & -\\
  & 166 & 1 & 1 & Y & & 1 & 1 & Y\\
  & 179 & 0.5 & 1 & Y & & - & - & -\\
  & 172 & 0.5 & 1 & Y & & - & - & -\\
  & 182 & 0.25 & 0.66 & N & & 0.25 & 0.67 & N\\
  & 185 & 1 & 1 & Y & & 0.5 & 1 & Y\\
  & 193 & 1 & 1 & N & & 1 & 1 & N\\
  & 205 & 1 & 1 & Y & & 1 & 1 & Y\\
  & 242 & 1 & 1 & Y & & 1 & 1 & Y\\
  \midrule
  \multicolumn{9}{c}{Y: Yes, N: No, *: No, does not express that there is no mutation in all of the available data} \\
  \multicolumn{9}{c}{for 2016. Only that the mutations are not present in the current random sampling.} \\
    \hline 
  \end{longtable}
\endgroup
\end{figure}